\documentclass[10pt,twocolumn,letterpaper]{article}

\usepackage[pagenumbers]{cvpr} 

\usepackage{graphicx}
\usepackage{amsmath}
\usepackage{amssymb}
\usepackage{booktabs}

\usepackage{amsfonts}
\usepackage{siunitx}
\usepackage{soul}
\usepackage[dvipsnames,table]{xcolor}
\usepackage{float}
\usepackage{parskip}
\usepackage[dvipsnames]{xcolor}
\usepackage{pifont}

\usepackage{comment}

\usepackage[pagebackref,breaklinks,colorlinks]{hyperref}
\usepackage{xspace}

\usepackage[capitalize]{cleveref}
\crefname{section}{Sec.}{Secs.}
\Crefname{section}{Section}{Sections}
\Crefname{table}{Table}{Tables}
\crefname{table}{Tab.}{Tabs.}

\newcommand{\D}{\mathcal{D}}

\newcommand{\R}{\mathbb{R}}
\newcommand{\Z}{\mathbf{z}}
\newcommand{\p}{\mathbf{p}}
\newcommand{\mem}{\text{mem}}
\newcommand{\m}{\mathbf{m}}

\newcommand{\cut}[1]{}
\renewcommand{\paragraph}[1]{\textbf{#1}}
\renewcommand{\th}{\textsuperscript{\,th}\xspace}

\DeclareMathOperator{\head}{head}
\DeclareMathOperator{\softmax}{softmax}

\def\cvprPaperID{8658} 
\def\confName{CVPR}
\def\confYear{2023}

\begin{document}

\title{À-la-carte Prompt Tuning (APT): Combining Distinct Data Via Composable Prompting}

\author{Benjamin Bowman$^{1,2}$\thanks{Work done during an internship at AWS AI Labs.}
\quad 
Alessandro Achille$^1$
\quad
Luca Zancato$^1$
\quad
Matthew Trager$^1$\\
Pramuditha Perera$^1$
\quad
Giovanni Paolini$^1$
\quad
Stefano Soatto$^1$\vspace{0.2em} \\
AWS AI Labs$^1$ \quad UCLA$^2$\vspace{0.2em}\\
{\tt benbowman314@math.ucla.edu \tt zancato@amazon.it}\\
{\tt\{aachille,mttrager,pramudi,paoling,soattos\}@amazon.com}
}
\maketitle

\begin{abstract}
We introduce À-la-carte Prompt Tuning (APT), a transformer-based scheme to tune prompts on distinct data so that they can be arbitrarily composed at inference time.  The individual prompts can be trained in isolation, possibly on different devices, at different times, and on different distributions or domains.  Furthermore each prompt only contains information about the subset of data it was exposed to during training.  During inference, models can be assembled based on arbitrary selections of data sources, which we call \textit{\`a-la-carte learning}. À-la-carte learning enables constructing bespoke models specific to each user's individual access rights and preferences.  
We can add or remove information from the model by simply adding or removing the corresponding prompts without retraining from scratch.  We demonstrate that \`a-la-carte built models achieve accuracy within $5\%$ of models trained on the union of the respective sources, with comparable cost in terms of training and inference time.  For the continual learning benchmarks Split CIFAR-100 and CORe50, we achieve state-of-the-art performance.
\end{abstract}

\section{Introduction}
\label{sec:intro}

As large neural network models make their way into commercial applications, the basic paradigm of training them on a monolithic dataset leads to a number of challenges. First, as new data become available, updating the whole model can be prohibitively expensive.  Even when training time is not an issue, some users may still require access and maintenance of previous versions of the model to avoid disruptions of their downstream workflows.  Second, owners of the training data may modify their sharing preferences at any time, leading to datasets that shrink over time (machine unlearning) or to different subsets of the training data being usable by different users (compartmentalization). Finally, the users themselves may want to use custom subsets of the data to better tailor their model to their use cases (model customization).

\begin{figure*}[t]
    \centering
    \includegraphics[width=0.95\linewidth,trim={0.5cm 3.5cm 0.5cm 4.5cm},clip]{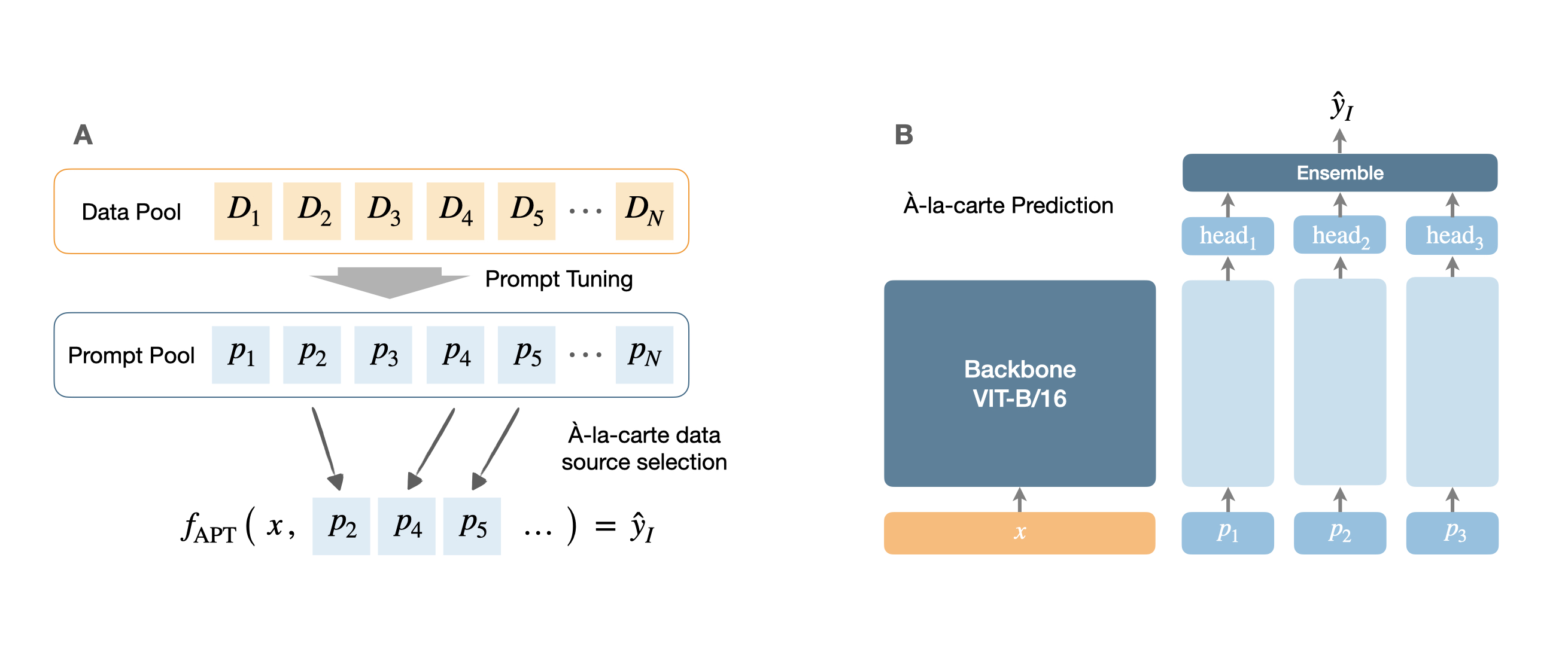}
    \caption{\textbf{À-la-carte Learning and APT.} Given a pool of multiple data sources, the goal of À-la-carte Learning is to allow the user to select -- at inference time -- an arbitrary subset $S \subset \D$ of sources to use. The performance of the \`a-la-carte model should be comparable to the performance of a model trained on $S$. \textbf{(A)} APT enables efficient À-la-carte Learning by converting each source into a prompt, and composing together the relevant prompts at inference time. \textbf{(B)} To perform inference, APT uses a modified attention mechanism that prevents the prompts from interfering with each other and ensembles the individual outputs to construct the final prediction.
    }
    \label{fig:a_la_carte_diag}
\end{figure*}

These challenges are well known and addressed separately in different fields such as continual learning, forgetting, and model adaption. However, in order for a commercial system to be viable at scale, these issues have to be tackled concurrently. Ideally, one would have a large model that each user can run, trained using only data the specific user wants and has rights to, that can evolve without the need for fine-tuning as new data becomes available, or as individual data owners exercise their right to have their data erased (``the right to be forgotten").

We refer to the problem of building such a model as \textit{\`a-la-carte learning} since, depending on the data availability and the user, the service may need to select and use different data chunks from a menu of available training data.  More specifically, let $\D = \{D_1, \ldots, D_n\}$ be a variable collection of data sources (a \textit{data pool}). In \`a-la-carte learning a user at inference time can specify a subset $S \subset \D$ of training data together with an input sample $x$ to receive a personalized \textit{\`a-la-carte} output $f(x, S)$ from the model $f$. 
Critically, the output $f(x, S)$ must not depend on any data source $D_i \notin S$.

À-la-carte learning can be na\"ively tackled in two ways. The service could pre-train one model for each possible subset of the data pool, and serve each user the most powerful model they have rights to. While optimal from the user view-point, this requires a prohibitive exponential complexity $O(2^{|\D|})$ in both training time and storage. On the other extreme, the service could train a separate model on each data source individually and, at inference time, ensemble all models obtained from the sources in $S$. This requires only linear $O(|\D|)$ training time complexity to pre-train each model, but still has a significant storage cost.  Furthermore due to the ensembling inference time is significantly increased while also potentially suffering from lower performance than the ideal ``paragon" model trained on the union of sources in $S$. The goal of  \`a-la-carte learning is to achieve performance as close as possible to the paragon without significantly increasing  inference or training time.

To address these key issues, we propose \textbf{À-la-carte Prompt Tuning (APT)}. 
APT leverages vision transformers and prompt tuning to solve the à-la-carte learning problem. First, APT converts each dataset $D_i$ into a learned prompt $p_i$, thus transforming the data pool into a \textit{prompt pool}. Then at inference time, given a subset of sources $S$ to use, APT retrieves all corresponding prompts and concatenates them together with the input. Surprisingly, we show that in most cases APT has performance comparable to the paragon of joint learning with all data in $S$. Moreover, since each prompt is trained on an individual dataset, information is naturally compartmentalized. Thanks to the small size of prompts and an efficient forwarding method, APT is significantly cheaper (in both storage and inference time) than ensembling models.

\begin{figure*}[t]
\centering
\begin{minipage}[t]{.3\linewidth}
\vspace{0pt}
\centering
\includegraphics[width=0.9\linewidth]{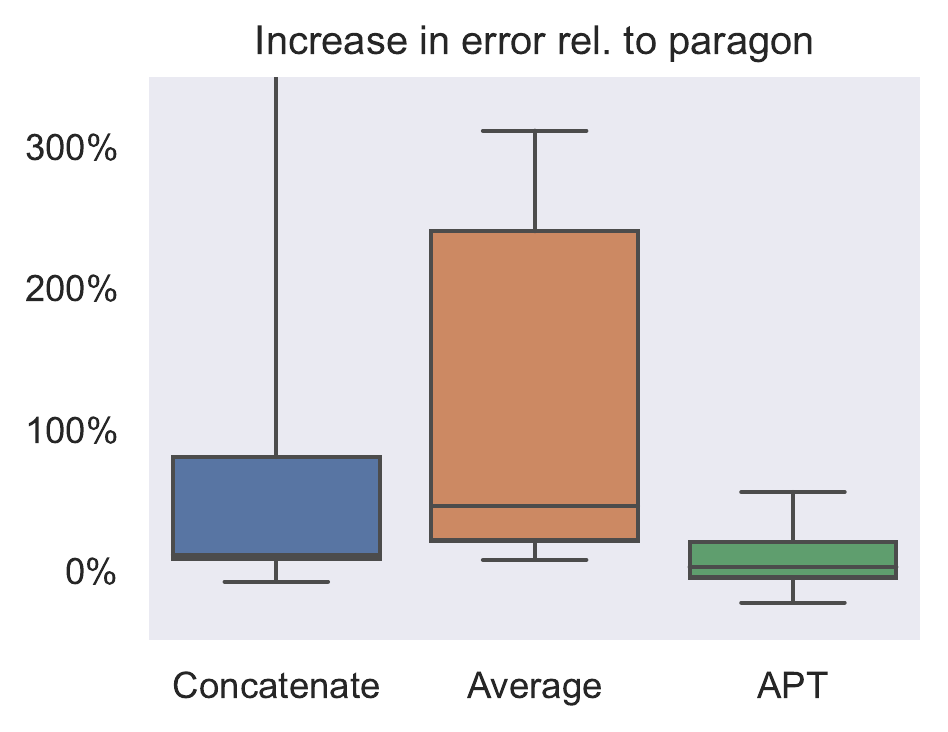}
\end{minipage}
\hspace{3em}
\begin{minipage}[t]{.4\linewidth}
\vspace{1.1em}
\centering
\colorlet{bbad}{red!80!black}
\colorlet{bad}{orange!85!black}
\resizebox{0.98\linewidth}{!}{
\begin{tabular}{c|ccc|c} \toprule
    {Dataset} & Concatenate & Average & APT & Paragon \\ \midrule
    MIT-67  & 84.6\% & 85.1\% & \textbf{86.2\%} & 86.2\%\\
    Cub-200 & \textcolor{bad}{85.2\%} & \textcolor{bad}{84.6\%} & \textbf{87.8\%} & 86.6\%  \\
    Caltech-256 & \textbf{91.1\%} & \textcolor{bad}{87.9\%} & \textbf{91.1\%} & 91.7\%  \\
    Pets  & \textbf{93.8\%} & 91.4\% & 93.1\% & 93.3\% \\ 
    Aircrafts & \textcolor{red!80!black}{56.5\%} & \textcolor{bbad}{16.7\%} & \textbf{61.1\%} & 71.0\%  \\
    Flowers & \textcolor{bbad}{84.5\%} & \textcolor{bad}{96.3\%} & \textbf{99.3\%} & 99.1\% \\
    Stanford Cars & \textcolor{bbad}{60.3\%} & \textcolor{bbad}{26.1\%} & \textbf{70.7\%} & 81.2\%
\end{tabular}
}
\end{minipage}
\caption{\textbf{Naive prompt composition vs. APT.} We compare different methods of combining prompts.  We split the training dataset into two equal sized shards then train prompts on each of the two shards in isolation.  We then compare the test accuracies after combining the prompts using different methods.  For the column ``Concat" we concatenate the prompts without structured attention and average ensemble their predictions.  For the column ``Avg" we simply average the prompts and classifier head as parameters and then take the single prediction.  The column ``APT" denotes our method.  Numbers more than $10\%$ below APT in each row are marked red; numbers more than $2\%$ below APT are marked orange.  The best method excluding the paragon in each row is marked in bold.}
\label{fig:naive-vs-cpt}
\end{figure*}

Importantly however, we note that simply concatenating different prompts that were trained separately leads to destructive interference in the attention block which corrupts the representations (see Table~\ref{fig:naive-vs-cpt}).  To address this problem, we introduce a modified attention mechanism that eliminates such interference, while also significantly reducing the inference time 
when multiple prompts are concatenated. A priori, this change comes with a small reduction in expressive power and in the ability to capture synergistic information between data sources. However, one of our main contributions is to show that the resulting drop in accuracy is generally modest, while providing far more valuable benefits to scalability, maintainability, and privacy.

We empirically demonstrate the advantage of APT-based \`a-la-carte learning for forgetting and continual learning (both domain-incremental and class-incremental). We observe that in most cases the performance of APT is within 5\% of the performance of the paragon at a fraction of the cost. We also show that APT outperforms all comparable baselines with the advantage of computational scalability from the structured attention mechanism.

\textbf{Summary of our contributions.}
\begin{enumerate}
    \item We introduce the À-la-carte Learning problem to address continual learning, machine unlearning, and model customization concurrently. 
    \item We propose APT, an efficient method to address À-la-carte Learning based on visual prompt tuning and a modified attention mechanism.
    \item We demonstrate that for most tasks APT achieves accuracy within $5\%$ of paragon performance even when each individual prompt has access to an order of magnitude less data
    \item We show that APT with a simple prompt weighting mechanism achieves state-of-the-art performance on continual learning benchmarks Split CIFAR-100 and CORe50.
\end{enumerate}


\section{Related Work}

\textbf{Prompt Tuning.}
Prompting originated from natural language processing by prepending ``hard'' language prompts to inputs to inform a pre-trained language model about the task to be solved \cite{liupretrainprompt, brown2020language}.  It was then discovered that one can optimize ``soft'' prompts in the embedding space in a differentiable fashion, with competitive performance to fine-tuning \cite{xianglisaliprefixtuning,lester-etal-2021-power, liu-etal-2022-p}.  This technique also proved useful when applied to vision transformers \cite{vpt_published}.  The idea of extending pre-trained transformers using prompt tokens with attention masking was introduced in \cite{learnablememory}.  We use the same attention masking scheme in our à-la-carte learning implementation.  The ensembling of soft and hard prompts was considered in \cite{lester-etal-2021-power} and \cite{clip} respectively.

\textbf{Continual Learning.}
Prompt tuning applied to the continual learning problem has been considered in \cite{dytox,continual_prompt_learning, wang2022sprompts}.  \cite{dytox} augment a fixed backbone with small task tokens that can be trained during episodes and added to the model incrementally.  In \cite{continual_prompt_learning} they query collections of prompts from a prompt pool on an instance-wise basis to be concatenated at inference time.  The query mechanism is supervised and consequently the compositionality of prompts is emergent from the supervision.  By contrast, we select prompts from a pool on a per-user basis and achieve composability of prompts through structured attention.  In \cite{wang2022sprompts} they address the domain incremental learning problem by training prompts independently on each domain and constructing a set of reference prototypes for the domain via $K$-means.  At inference time, given an input $x$ they select the prompt according to the closest reference prototype to the embedding of the point $x$.  In our APT Weight (APT-W) scheme (described in \cref{par:class_incremental_learning}) rather than select a single prompt we weight the prompts according to the instance embedding's distance to the closest prototype.

\textbf{Forgetting.} Forgetting in deep networks \cite{Golatkar_2020_CVPR,golatkar2020forgetting} is challenging. \cite{mixedprivacyforgettinggolatkar} utilizes a ResNet-50 where they train a linearization of the network starting from a pre-trained checkpoint.  Due to the linear parameterization, forgetting is much more tractable and they can get a bound on the mutual information after a certain number of forgetting steps.  \cite{approximatedatadeletion} offers forgetting for linear/logistic models, and \cite{descenttodelete} offer forgetting techniques in the convex setting.  \cite{machineunlearning} investigated training distinct networks on separate shards of data.  We run this same procedure to benchmark our APT approach.  The novelty with the prompt tuning approach is that the memory overhead is minimal, and inference can be done at the cost of a single forward pass. 

\section{Preliminaries}

\paragraph{Vision Transformer.} We use vision transformers \cite{dosovitskiy2021an} as our backbone architecture, due to both good accuracy on downstream tasks and ease of prompting. An image $x \in \mathbb{R}^{H \times W \times C}$ is split into $N$ patches $x^{(1)}, \ldots, x^{(N)}$, which are represented as $d$-dimensional tokens $z^{(i)} = E x^{(i)} + e_\text{pos}^{(i)} \in \R^d$ through a learned linear embedding $E$ and a set of positional encodings $\{e_\text{pos}^{(i)}\}_{i=1}^N$.  We add a special learnable \textit{class} token $z^{(0)}$ that is shared by all inputs.  The input to the first layer of the transformer is then given by $\Z_0 := [z^{(0)}, z^{(1)}, \ldots, z^{(N)}]$ which is the concatenation of the class token and the tokens corresponding to the image patches.  Let $F^\ell_\theta$ denote the $\ell$\th attention layer of the transformer, where $\theta$ denotes the parameters of the model. The output tokens of the $\ell$\th layer are given by
\[ \Z_\ell :=  F^\ell_\theta(\Z_{\ell - 1}). \]
Let $z_L^{(0)}$ be the output of the class token at the last transformer layer.
We use a linear head to output a probability distribution $\hat{y}$ of the input's label:
\[
\hat{y} := \softmax(\head_\theta(z_L^{(0)}))
\]
where $\head_\theta(x) = W x + b$ is a learned fully connected layer.

\paragraph{Visual Prompting.} Like convolutional networks, pre-trained vision transformers can be adapted to new downstream tasks by fine-tuning their weights $\theta$. However, prompting can also be used as an alternative adaptation mechanism for vision transformers \cite{vpt_published,learnablememory}.
Let $D$ be a supervised dataset for a downstream task. A new learnable \textit{prompt token} $p_0$ is attached to the transformer's input, so that
the final output is given by
\[
[\Z_L, p_L] = F_\theta^L \circ \ldots \circ F_\theta^1 ([\Z_0, p_0]).
\]
To predict the downstream task label, the head of the pre-trained model is discarded to be replaced by a new head which is trained on the final prompt token
\[
\hat{y} = \softmax(\head(p_L)).
\]
Both $p_0$ and $\head$ are trained on $D$, while the parameters $\theta$ of the pre-trained backbone are frozen.

\paragraph{Notation.} We denote with $\ell(\hat{y}, y)$ the cross entropy loss, and for a natural number $k \in \mathbb{N}$ we let $[k] := \{1, \ldots, k\}$.  We consider a classification task where $\mathcal{X}$ is the input domain and $\mathcal{Y}$ is the label space.  

\begin{figure*}
    \centering
    \includegraphics[width=0.95\linewidth]{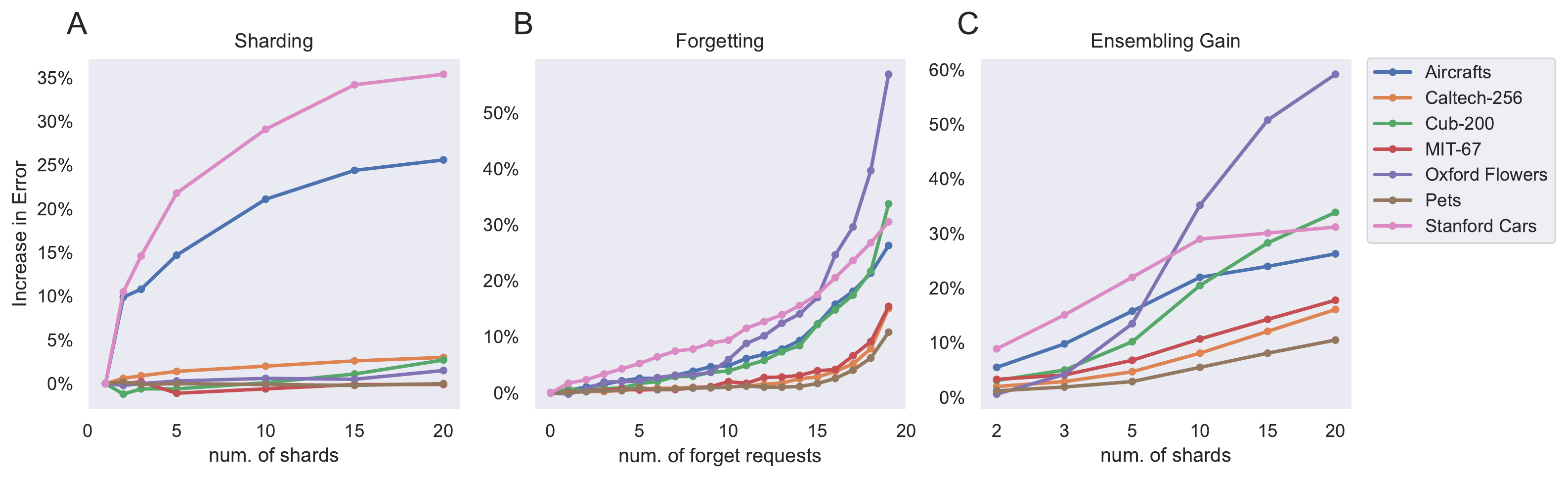}
    \caption{
    \textbf{(A) Error increase of APT compared to paragon.}
    We split a training set into a varying number of equal sized shards chosen uniformly at random. We then use APT to combine prompts learned individually on each shard, and measure the increase in error compared to the paragon of training on all data together.  For most datasets, the performance of the APT is within a few percent of the paragon, even when the dataset is split in up to 20 parts. \textit{Aircrafts} and \textit{Stanford Cars} are the main exceptions, possibly due to the large domain shift between the backbone pretraining and those tasks. \textbf{(B) Satisfying forgetting requests.} We simulate a sequence of data removal requests starting from a pool of 20 sources and removing one source at the time. We report the increase in error compared to using the full data. We see that APT degrades gracefully as desired, while also ensuring perfect data removal. 
    \textbf{(C) Gain of using ensembles instead of individual prompts.} We split a train set in a varying number of shards, and show the difference between the accuracy of APT prompt composition and the average accuracy of the individual prompts. For large number of shards, individual prompts don't have enough information to classify accurately but APT can combine them to create a much stronger classifier (with up to 60\% better accuracy).
    }
    \label{fig:acc_vs_shards}
\end{figure*}

\section{À-la-carte Prompt Tuning}

Suppose we have a pre-trained backbone $f_\theta$ and a pool of additional data sources $\mathcal{D} := \{D_1, \ldots, D_n\}$. We focus  in particular on the case where all sources in $\D$ pertain to the same task and share the input and label space $D_i \subset \mathcal{X} \times \mathcal{Y}$.\footnote{We do not however need to assume that all sources contain samples from all classes. The backbone $f_\theta$ may be pre-trained on the same task as $\D$ (in which case $\D$ provides additional data to tune the model) or may be pre-trained on an unrelated proxy task (e.g., ImageNet or web-scale data).}
Ideally we would like to fine-tune the backbone using all data in $\D$ by minimizing the loss:
\[
L_\D(\theta) = \sum_{(x, y) \in \bigcup \D} \ell (y, f_\theta (x)).
\]
However, it is often the case (see \Cref{sec:applications}) that the collection of data sources $\D$ changes over time as data is added or removed. It may also be the case that different users of the model may want to use different subsets of the data to better cover their use cases (model customization) or may only have access rights to certain subsets of the data (compartmentalization).

\textbf{À-la-carte Learning.} To remedy this, \textit{at inference time}, given any subset $I \subset [n]$ we would like to be able to use a model that uses data exclusively from $\mathcal{D}_{I} := \bigcup_{i \in I} D_i$. A trivial option is to fine-tune in advance the parameters $\theta_I$ on each possible subset $I$ minimizing the loss
\[ L_{\D_{I}}(\theta_I) := \sum_{(x, y) \in \D_{I}} \ell(f(x; \theta_I), y) \]
and, given $I$ at inference, select the corresponding $\theta_I$ and use it to form the model $f(x;\theta_I)$. However, since there are $2^n$ possible subsets $I \subset [n]$ it is prohibitively expensive to fine-tune a separate model for each $I$, both from a compute-time and storage cost perspective. It would also require training $2^n$ new models each time a source of data is added, which becomes infeasible quickly.

\textbf{Na\"ive À-la-carte Prompt Tuning.} To reduce the computational cost while satisfying all requirements of À-la-carte Learning, we suggest an alternative strategy based on composition of prompts trained on individual data sources.
For each $i \in [n]$ we train a prompt $p_i$ and classifier head $\mathrm{head}_i$ on the data $D_i$ using the loss function
\[ L_{D_i}(p^{(i)}, \mathrm{head}_i) := \sum_{(x, y) \in D_i} \ell(f(x; p^{(i)}), y) \]
where the dependence of $f(x; p^{(i)})$ on $\mathrm{head}_i$ above has been suppressed for ease of notation.  Given a set of indices $I=\{i_1, \ldots, i_{|I|}\}$ we denote with $\p^{(I)} = [p^{(i_1)}, \ldots, p^{(i_{|I|})}]$ the concatenation of all prompt tokens corresponding to each data source in $\mathcal{D}_I$.  The final output of the transformer is given by
\[
[\Z_L, \p_L^{(I)}] := F_\theta^L \circ \ldots \circ F_\theta^1 ([\Z_0, \p^{(I)}])
\]
where $\theta$ are the frozen parameters of the backbone transformer.  Each output token $p_L^{(i)}$ corresponding to a prompt $p^{(i)}$ can be used to generate a prediction
\[
\hat{y}^{(i)} := \softmax(\head_i(p_L^{(i)})).
\]
The final prediction is made by  ensembling the predictions made by each individual prompt $p^{(i)}$ (see also \Cref{fig:a_la_carte_diag}):
\[
\hat{y}_I := \frac{1}{|I|} \sum_{i \in I} \hat{y}^{(i)}.
\]
Since each prompt only contains information about its own source, the model output $\hat{y}_I$ depends only on the sources in $\D_I$. Moreover, after the initial cost $O(|\D|)$ to train each prompt $p_i$, any subset $I$ of sources can be combined at inference time with constant cost $O(1)$. Hence, this procedure satisfies the requirements for à-la-carte learning.

However, in \Cref{fig:naive-vs-cpt} we see that the performance of this na\"ive implementation of \`a-la-carte prompt tuning by concatenating prompts severely underperforms the paragon of using a single prompt trained from scratch on all the datasets in $\D_I$. The same is true for other composition mechanisms, such as averaging prompts. We hypothesise that this is due to the prompts, which were trained individually, corrupting the representations at inference time when concatenated due to destructive interference in the attention mechanism of the transformer.

\paragraph{Structured Attention.}
To remedy this we follow the technique in \cite{learnablememory}. First, we mask the attention so that the $\Z_\ell$ tokens do not attend to the prompts, and the prompts do not attend each other (see \Cref{fig:attn_mask_tab}). This ensures the result of forwarding each prompt $p^{(i)}$ through the network is unaffected by the presence of the other prompts. However, this reduces the power of the prompts to modify the forward pass of the network. To compensate, at each layer $\ell$ of the transformer and for each prompt $p^{(i)}$ we add a set of $d_\mem$ learnable memory tokens $\m_\ell^{(i)} \in \R^{d_\mem \times d}$. These memory tokens can be attended by the prompts but cannot attend to anything. While a similar result could be obtained by using longer prompts instead of using memory tokens, \cite{learnablememory} notes that this solution gives comparable accuracy with a significantly reduced inference time.  Due to the structured attention, a single forward pass of the backbone transformer can be performed on $\Z_0$ independent of the prompts.  Subsequently at each layer $l$ each prompt $p_{\ell}^{(i)}$ can perform cross attention to query itself and $[\Z_{\ell}, \m_{\ell}^{(i)}]$.  While self-attention has quadratic complexity in the sequence length, this implementation has $O(N^2 + (N + d_{mem})|I|)$ complexity as opposed to $O((N + |I|)^2)$ complexity for self-attention without memory.  Consistent with \cite{learnablememory} in our implementation we set $d_{mem} = 5$.  Consequently $N^2 \gg (N + d_{mem})$, and thus adding a prompt, and thus increasing $|I|$, only marginally increases inference time relative to the fixed cost of a forward pass for the backbone transformer $O(N^2)$.  By contrast classic model ensembling would have inference cost $O(|I| N^2)$ as one must do a forward pass through each model.  Furthermore each prompt corresponds to $12 \times d_{mem} + 1$ tokens, which amounts to a number of parameters less than $.06\%$ of the backbone model.  Thus the memory overhead of storing the prompts is also marginal.

\definecolor{yes}{RGB}{142,195,225}
\definecolor{no}{RGB}{223,235,243}
\newcommand{\cmark}{\ding{51}}%
\newcommand{\xmark}{\ding{55}}%
\newcommand{\y}{\cellcolor{yes}\cmark}
\newcommand{\n}{\cellcolor{no}\xmark}
\begin{figure}[h]
    \centering
    \resizebox{0.8\columnwidth}{!}{%
    \begin{tabular}{c|c|c|c|c|c|c|c}
        & $\mathbf{z}_\ell$ & $p^{(1)}_\ell$ & $p^{(2)}_\ell$ & $p^{(3)}_\ell$ & $\mathbf{m}^{(1)}_\ell$ & $\mathbf{m}^{(2)}_\ell$ & $\mathbf{m}^{(3)}_\ell$ \\
      \hline
      $\mathbf{z}_\ell$ & \y & \n & \n & \n & \n & \n & \n \\
      $p^{(1)}_\ell$ & \y & \y & \n & \n & \y & \n & \n \\
      $p^{(2)}_\ell$ & \y & \n & \y & \n & \n & \y & \n \\
      $p^{(3)}_\ell$ & \y & \n & \n & \y & \n & \n & \y
    \end{tabular}%
    }
    \caption{\textbf{Attention Masking Table.}  The rows correspond to queries and the columns correspond to keys.  The cells marked with \cmark\, denote where attention is performed and the cells marked \xmark\, denote where attention is masked.}
    \label{fig:attn_mask_tab}
\end{figure}

\textbf{À-la-carte Prompt Tuning.} Our final proposal for efficient À-la-carte Learning, which we call \`A-la-carte Prompt Tuning (APT), combines the composition of individual prompts with the structured attention mechanism. In \Cref{fig:naive-vs-cpt} we see that APT outperforms the na\"ive baselines in almost all cases, and importantly it not prone to the same catastrophic failures (e.g. on Aircrafts and Stanford Cars). Moreover its performance is close or better\footnote{The results better than the paragon can be attributed to the regularization effect of ensembling prompts trained on different subsets of the data.} than the paragon performance (training a prompt directly on the union of all datasets) on all datasets except Aircrafts and Stanford Cars. In the following, we explore particularly interesting applications of À-la-carte learning, and we empirically test the performance of APT in different settings.

\begin{table*}\small
\centering
\begin{tabular}{c|c|cccccc} \toprule
    {Dataset} & {No Sharding} & {2 Shards} & {3 Shards} & {5 Shards} & {10 Shards} & {15 Shards} & {20 Shards}  \\ \midrule
    Head-only (in-domain) & 90.8\% & 90.8\% & 90.8\% & 90.4\% & 90.1\% & 89.5\% & 88.5\% \\
    APT (in-domain) & \textbf{91.4}\% & \textbf{91.5}\% & \textbf{91.3}\% & \textbf{91.4}\% & \textbf{91.0}\% & \textbf{90.6}\% & \textbf{90.0}\% \\ \midrule
    Head-only (out-of-domain) & 59.7\% & 56.5\% & 53.5\% & 50.5\% & 45.0\% & 41.6\% & 40.5\% \\
    APT (out-of-domain) & \textbf{76.1}\% & \textbf{65.9}\% & \textbf{63.4}\% & \textbf{57.9}\% & \textbf{51.0}\% & \textbf{46.8}\% & \textbf{45.6}\% 
\end{tabular}
\caption{\textbf{Head-only ensembling vs. APT.} We compare the performance of APT to ensembling classifier heads (without prompts) trained on distinct shards chosen uniformly at random.  We group the datasets MIT-67, Cub-200, Caltech-256, Pets, and Flowers as ``in-domain" due to their alignment with ImageNet21k and group the datasets Aircrafts and Stanford Cars as ``out-of-domain" due to their difference with the pretraining.  We report the average accuracy for the datasets within each group.  We see that APT consistently outperforms Head-only ensembling, and the difference is most pronounced for out-of-domain datasets.}
\label{table:apt_vs_head_only}
\end{table*}

\section{Applications of À-la-carte Learning}\label{sec:applications}
\textbf{Decentralized Learning.} We may have datasets $D_1, \ldots, D_n$ stored across $n$ different servers or devices.  Each server can train a prompt $p_i$ on $D_i$ in isolation.  At inference time, we can assemble the prompts $p_1, \ldots, p_n$ on a central server and perform inference using $\p_{[n]}$.  Each server can train their prompt $p_i$ without exposing or uploading their raw data to the central server.  This is useful whenever it is not possible to efficiently aggregate the data across the different devices, or if the individual devices are not willing to expose their raw data.  We note that in Federated learning one typically looks at a different setting where a single central model is trained but the gradients are computed locally and then shared.  Since a single model is trained via gradients aggregated across all the sources, this does not solve the \`a-la-carte learning problem and does not allow forgetting a source of data or firewalling a particular user from a source of data. Nevertheless, the two approaches are not mutually exclusive and we believe integrating them is an interesting avenue of research.

\textbf{Model Versioning.} Different users may have different rights in terms of which datasets they are permitted to access.  For each user $A$ we can associate a set of indices $I \subset [n]$ based on which datasets they have rights to.  Then the version of the model we offer to user $A$ would be given by $f(x; \theta_I)$.  Aside from dataset rights, individuals may wish to add or drop data from the influence of a model simply for performance reasons. A dataset $D_i$ may be useful for user $A$ but introduce performance degradations for user $B$.  \`A-la-carte learning allows us to include or not include the prompt $\theta_i$ for different users.  Furthermore since the prompts do not need to be trained at the same time, we can add prompts at later points in time to update the model according to new data.

\par \textbf{Forgetting.} Forgetting a source $D_i$ is easy, as we simply need to delete its associated prompt $p_i$.  However, a service may periodically get requests to forget specific samples $(x, y)$.  Retraining a model from scratch each time a forget request is received can be prohibitively expensive.  Furthermore even if the economic cost of training is no issue, satisfying the forget request immediately requires suspending the service until the retraining has completed which can induce service delays.  Following \cite{machineunlearning}, we can partition our dataset $\mathcal{D}$ into disjoint ``shards" of equal size chosen uniformly at random so that $\mathcal{D} = \bigcup_{i \in [n]} D_i$.  Then anytime we receive a request to forget a specific data point $(x, y) \in \D$ we only need to retrain the prompt $p_i$ corresponding to the shard $\D_i$ that $(x, y)$ belongs to.  Furthermore the forget request can be satisfied immediately without any downtime to the service as the service can drop the prompt $p_i$ from the model while it is being retrained and form predictions using the remaining prompts in the meantime.

\paragraph{Continual Learning.} We can let $D_i$ each correspond to a different training episode.  Then in a continual learning setting at each episode $i$ we train a prompt $p_i$ on $D_i$ and let our model after the specific training episode be $f(x; \p_I)$ where $I = \{1, 2, \ldots, i\}$.

\section{Experiments}
\label{sec:experiments}

In all experiments we use a VIT-B/16 \cite{dosovitskiy2021an} pre-trained on ImageNet-21k.  Unless explicitly stated otherwise, we use the pre-trained model \texttt{vit\_base\_patch16\_384} from the timm\footnote{\url{https://github.com/rwightman/pytorch-image-models}} library in PyTorch \cite{pytorch}. 

\paragraph{Datasets.} We evaluate APT on the datasets MIT-67 \cite{mit67recognizing}, Cub-200-2011 \cite{WahCUB_200_2011}, FGVC-Aircrafts \cite{maji13fine-grained}, Oxford Flowers \cite{Nilsback06}, Caltech-256 \cite{griffin_holub_perona_2022}, Oxford Pets \cite{parkhi12a}, and Stanford Cars \cite{KrauseStarkDengFei-Fei_3DRR2013}.  Based on the distance from the ImageNet21k pre-training, similarly to \cite{Li2020Rethinking} we classify the datasets MIT-67, Cub-200-2011, Oxford Flowers, Caltech-256, and Oxford Pets as ``in-domain" datasets and classify the datasets FGVC-Aircrafts and Stanford Cars as ``out-of-domain" datasets.  To test APT on class incremental learning problem we use Split CIFAR-100 \cite{krizhevsky2009learning} (10 training episodes, 10 classes per episode) and for domain incremental learning we use CORe50 (8 training domains, 3 test domains) \cite{pmlr-v78-lomonaco17a,Lomonaco2019FineGrainedCL}.

\begin{table*}[t]\small
\centering
\begin{tabular}{c|c|ccccccc} \toprule
    {Dataset} & {No Sharding} & {2 Shards} & {3 Shards} & {5 Shards} & {10 Shards} & {15 Shards} & {20 Shards} & {50 Shards} \\ \midrule
    MIT-67  & 86.2\% & 86.2\% & 86.0\% & 87.3\% & 86.8\% & 86.3\% & 86.3\% & 84.2\% \\
    Cub-200  & 86.6\%  & 87.8\% & 87.2\% & 87.2\% & 86.5\% & 85.5\% & 83.9\% & 79.9\%  \\
    Caltech-256  & 91.7\%  & 91.1\% & 90.8\% & 90.3\% & 89.7\% & 89.1\%  & 88.7\% & 87.1\% \\
    Pets  & 93.3\%  & 93.1\% & 93.4\% & 93.3\% & 93.4\% & 93.5\% & 93.3\% & 92.3\% \\ 
    Aircrafts & 71.0\% & 61.1\% & 60.2\% & 56.3\% & 49.9\% & 46.6\% & 45.4\% & 36.9\%\\
    Flowers & 99.1\% & 99.3\% & 99.1\% & 98.8\% & 98.5\% & 98.6\% & 97.6\% & 97.7\% \\
    Stanford Cars & 81.2\% & 70.7\% & 66.6\% & 59.4\% & 52.1\% & 47.0\% & 45.8\% & 39.1\% \\ \midrule
     Average & 87.0\% & 84.2\% & 83.3\% & 81.8\% & 79.6\% & 78.1\% & 77.3\% & 73.9\%
\end{tabular}
\caption{\textbf{Accuracy of shard ensembles.} Accuracy of ensembling prompts trained on disjoint shards chosen uniformly at random. We
see that for many datasets the performance of the ensemble is close to the paragon of prompt tuning on the entire dataset, despite each
predictor of the dataset only seeing a fraction of the entire dataset.}
\label{table:shard_accs}
\end{table*}
\begin{table*}[ht]\small
\centering
\begin{tabular}{c|c|cc|ccc|c} \toprule
    {Dataset} & {Finetuning} & {Head-only} & {Bias+Head} & {Deep PT} & {Deep Shared PT} & {Shallow PT} & {FT vs. PT gap} \\ \midrule
    MIT-67 & 87.1\% & 85.6\% & 87.2\% & 86.2\% & 86.5\% & 86.0\% & -0.9\% \\
    Cub-200 & 88.4\%  & 87.0\% & 89.4\% & 86.6\% & 86.4\% & 85.6\% & -1.8\% \\
    Caltech-256 & 93.5\% & 90.4\% & 93.0\% & 91.7\% & 91.3\% & 90.4\% & -1.8\% \\
    Pets & 94.5\% & 92.2\% & 94.9\% & 93.3\% & 92.9\% & 91.6\% & -1.2\%     \\ 
    Aircrafts & 75.6\% & 54.8\% & 75.6\% & 71.0\% & 68.2\% & 62.1\% & -4.6\% \\
    Flowers & 97.4\% & 98.8\% & 99.4\% & 99.1\% & 98.9\% & 98.5\% & 1.7\%  \\
    Stanford Cars & 84.3\% & 64.5\% & 86.6\% & 81.2\% & 78.6\% & 69.6\% & -3.1\%  \\ \midrule
    Avg & 88.7\% & 81.9\% & 89.4\% & 87.0\% & 86.1\% & 83.4\% & -1.7\%
\end{tabular}
\caption{\textbf{Finetuning vs. Prompt Tuning.} We compare different finetuning methods to prompt tuning.  In the ``Head-only" column only the linear classifier head is trained.  In ``Bias+Head" the bias's as well as the classifier head are trained.  ``Deep PT" is prompt tuning with memory tokens at each layer.  ``Deep Shared PT" is prompt tuning where the memory tokens are shared across the layers.  In ``Shallow PT" a single prompt is tuned without memory tokens.  ``FT vs. PT Gap" reports the accuracy of Deep PT minus the accuracy of Finetuning.}
\label{table:pt-vs-ft}
\end{table*}

\textbf{Comparison of model-tuning methods.}
Since our method is based on prompt-tuning, in Table~\ref{table:pt-vs-ft} we measure how it compares to standard fine-tuning. Consistent with \cite{vpt_published}, we see that on most datasets prompt tuning is competitive (within $2\%$ accuracy) with finetuning and outperforms head-only tuning, especially on out-of-domain datasets. We also observe that per-layer memory tokens (Deep PT) have the best trade-off between accuracy and computational cost, motivating our design choice to use it.

\textbf{Decrease in performance due to sharding.}
Given a sharded dataset, we aim to establish whether composing per-shard prompts using APT achieves a comparable performance to training a prompt on all available data (paragon).
Following \cite{machineunlearning}, we split the training set into disjoint shards of equal size.  The splitting is done by selecting samples uniformly at random, hence the number of examples per class can slightly vary across shards and smaller shards may not have examples from all classes.  We train prompts on each of the shards in isolation and then compose a model using APT. The test accuracies as we increase the number of splits are reported in Table~\ref{table:shard_accs}. Figure~\ref{fig:acc_vs_shards} (A) shows the increase in test error of the APT method relative to the paragon.  As expected, the accuracy of APT generally decreases as the number of splits increases. However, for many datasets the drop off in accuracy is surprisingly small: on the in-domain datasets for 10-20 shards the accuracy of APT is within 2-5\% of the accuracy of the paragon of training on the entire dataset.  The main exceptions are out-of-domain datasets, where we observe a steeper accuracy drop when splitting the dataset. We hypothesize that for out-of-domain dataset, synergistic information between datapoints of different shards is more important for the training process.
\par

\textbf{Importance of composing prompts.}  In Figure~\ref{fig:acc_vs_shards} (C) we plot the gap between the average individual prompt accuracy and the accuracy of APT.  We see that as the number of shards increases, the difference grows.  This implies that while the performance of the ensemble may drop off slowly, that the performance of the individual predictors is deteriorating.  This demonstrates that on large and fragmented data pools, individual prompts do not have enough information to classify correctly, and aggregating their information through the APT composition mechansim is essential.

\textbf{Ablations.}
We perform a series of ablations to piece out the essential components of APT.  To understand the effect of the attention masking, in \Cref{fig:naive-vs-cpt} we compare APT to the na\"ive method of concatenating all prompts without structured attention.  We see that naive concatenation performs almost uniformly worse than APT on average and has significantly higher variance, failing with very low accuracies in some cases.  To isolate the effect of prompt tuning on the success of APT, in Table~\ref{table:apt_vs_head_only} we compare our APT method to training a simple head-only classifier on each shard.  We see that APT uniformly outperforms its head-only counterpart, and that the difference is especially pronounced for out-of-domain datasets.

\paragraph{Forgetting sources.}
In Figure~\ref{fig:acc_vs_shards} (B) we plot the increase in error of the APT method after a certain number of shards (and their corresponding prompts) are deleted.  This simulates a setting where a service provider receives a sequence of forget requests and consequently must remove prompts from the model.  We see that starting with $20$ shards, even after removing $10$ shards for most the datasets the decline in accuracy is approximately $5\%$ or less despite utilizing half the data.  
Since training time is directly proportional to the number of samples in the training set, this implies that we can reduce the cost of retraining after a forget request by an order of magnitude with a negligible drop in accuracy.  Furthermore as shown in Figure~\ref{fig:acc_vs_shards} (B) we can handle a large number of forget requests sequentially without retraining before accuracy meaningfully declines.
Moreover, since adding and removing sources are symmetric operations for APT, the same plot can be interpreted in reverse as showing the performance of APT in incrementally learning from an increasing set of data sources.

\begin{table}[t]\small
\centering
\begin{tabular}{c|c|c} \toprule
    {Method} & {CIFAR-100} & CORe50\\ \midrule
    APT & 83.63 & 90.89 \\
    APT-W & \textbf{85.21} & \textbf{91.14}\\
    \midrule
    L2P \cite{l2p-short} & 83.83 & 78.33\\
    S-iPrompts \cite{s-prompts-short} & N/A & 83.13\\
    S-liPrompts \cite{s-prompts-short} & N/A & 89.06\\
    LwF \cite{lwf-short} & 60.69 & 75.45\\
    EWC \cite{ewc-short} & 47.01 & 74.82
 \end{tabular}
\caption{\textbf{Performance on Split CIFAR-100 and CORe50.} Reporting average accuracy on the test set.  Numbers for the non-APT methods are reported in \cite{s-prompts-short} or \cite{l2p-short}.  For fair comparison against \cite{l2p-short, s-prompts-short} we have changed the resolution of our VIT to 224 from 384.  Since APT does not train with a memory buffer we compare against the memoryless versions of L2P and S-Prompts.}
\label{fig:continuallearningeval}
\end{table}

\paragraph{Class Incremental Learning.}\label{par:class_incremental_learning}
Oftentimes one wishes to add new classes to the model incrementally.  In this section we explore class-incremental-learning (CIL) where at different training episodes we have access to different classes.  To evaluate APT in this setting, we use the Split CIFAR-100 benchmark where the dataset is split into 10 disjoint sets of classes, with each subset containing 10 classes each.  We train prompts on each subset in isolation.  At inference time, we simply concatenate the class predictions from the individual prompts in line with our APT scheme.  In \cref{fig:continuallearningeval} we report the results of APT in this setting.  Out-of-the-box APT outperforms all the baselines and has a comparable performance to L2P \cite{l2p-short}.  We note that an advantage of L2P is the ability to dynamically select the prompts based on the test sample. Since prompts in APT are compositional by construction, we can easily implement a similar mechanism.  Similarly to \cite{wang2022sprompts} we perform $K$-means on each episode in the embedding space to extract reference prototypes for that episode ($K=20$), then at inference time we weight each episode prompt based on the distance of the instance's embedding from that episode's prototypes. See details in \cref{sec:apt-weight} in the supplementary material for the exact weighting scheme.  We denote this method in the tables as APT-Weight (APT-W), and note that using this hard-coded weighting strategy -- in contrast with L2P's learned prompt selection mechanism -- APT-W outperforms L2P.  We note that this weighting scheme still satisfies the \`a-la-carte learning requirement since the reference prototypes for each source are constructed independently of the other sources.

\paragraph{Domain Incremental Learning.}
Oftentimes one encounters data from different domains at different points in time.  In the continual learning literature this setting is referred to as domain-incremental-learning (DIL).  In this section we evaluate APT on domain incremental learning.  In \cref{fig:continuallearningeval} we report the results of running APT on the CORe50 domain incremental learning benchmark.  The CORe50 dataset contains data from 8 training domains and 3 test domains.  By training prompts independently on each of the training domains, out-of-the-box APT outperforms all other methods on CORe50.  Weighting the prompts in the APT-W scheme seems to give only a marginal increase (0.25\%) in performance.

\section{Conclusion}
We introduced the general problem of \textit{\`A-la-carte Learning} and an efficient solution to the problem using \`A-la-carte Prompt Tuning (APT).  We demonstrate that models constructed \`a la carte are competitive with models trained on the union of the respective sources, with added benefits for privacy and customization.  Furthermore APT achieves state-of-the-art performance for both class incremental learning and domain incremental learning with additional benefits for privacy and customization.  While APT offers one solution to the \`A-la-carte Learning problem, we emphasize that this problem is more general and deserves further study in order to develop competitive machine learning methods that respect users' data usage and privacy rights.


{\small
\bibliographystyle{ieee_fullname}
\bibliography{egbib}
}

\clearpage

\renewcommand{\thesection}{\Alph{section}}
\renewcommand{\thesubsection}{\Alph{section}.\arabic{subsection}}
\setcounter{section}{0}

\section*{Supplementary Material}
\section{Details of APT Weight (APT-W)}\label{sec:apt-weight}
In this section we describe the details of the APT Weight (APT-W) scheme.  Let $\D = \{D_1, \ldots, D_n\}$ be a collection of sources.  Consistent with APT for each source $D_i$ we train a prompt $p^{(i)}$ and a classifier head $\head_i$ using only the data in $D_i$.  Then, differing with classical APT, for each source $D_i$ we perform $K$-means ($K=20$) in the embedding space to construct a set of prototypes $\mu_1^{(i)}, \ldots, \mu_K^{(i)}$.  More concretely for each $(x, y) \in D_i$ we forward the input $x$ through the transformer to get the final embedding sequence $[\Z_L(x)]$.  We use the class token embeddings $\Z_L^{(0)}(x)$ as the vectors to perform the $K$-means algorithm on.  Specifically we perform $K$-means on the set
\[ \{ \Z_L^{(0)}(x) : (x, y) \in D_i\} \]
to construct the prototypes $\mu_1^{(i)}, \ldots, \mu_K^{(i)}$.  At inference time, the basic forward pass for APT Weight is the same as APT.  Given an instance $x$ and a set $I = \{i_1, \ldots, i_{|I|}\} \subset [n]$ we let $\p^{(I)} = [p^{(i_1)}, \ldots, p^{(i_{|I|})}]$ be the concatenation of all prompt tokens corresponding to each data source in $\mathcal{D}_I$.  The final output of the transformer is given by
\[
[\Z_L, \p_L^{(I)}] := F_\theta^L \circ \ldots \circ F_\theta^1 ([\Z_0, \p^{(I)}])
\]
where the structured attention is applied as usual.  Each output token $p_L^{(i)}$ corresponding to a prompt $p^{(i)}$ is used to generate logits
\[
\hat{y}^{(i)} := \head_i(p_L^{(i)}).
\]
In contrast to APT, APT Weight will apply a weighting to the logits $\hat{y}^{(i)}$ based on the distance of the embedding of the instance $x$ to the prototypes $\mu_1^{(i)}, \ldots, \mu_K^{(i)}$.  Specifically, for each index in $i \in I$ we compute
\[ d_{i} = \min_{k \in [K]} \|\Z_L^{(0)} - \mu_k^{(i)}\|_2. \]
Let us denote $\mathbf{d} = (d_{i_1}, \ldots, d_{i_{|I|}})$.
Then we construct a weight vector
\[ w := \softmax(-\beta \mathbf{d}) \]
where $\beta$ is the inverse temperature which in our experiments we set to $\beta=0.1$.  We then form the weighted logits
\[ [w_{i_1} \cdot \hat{y}^{(i_1)}, w_{i_2} \cdot \hat{y}^{(i_2)}, \ldots, w_{i_{|I|}} \cdot \hat{y}^{(i_{|I|})}]. \]
For class incremental learning problems, these weighted logits are the final logits used for prediction.  For domain incremental learning problems, the logits are average pooled to form the final logits
\[ \hat{y} = \frac{1}{|I|} \sum_{i \in I} w_{i} \cdot \hat{y}^{(i)}. \]
\section{Hyperparameters}
Consistent with \cite{continual_prompt_learning} for the continual learning experiments on Split CIFAR-100 and CORe50 we train for 5 epochs.  All methods in Table~\ref{table:pt-vs-ft} are trained for 150 epochs.  For all other experiments the paragon method (trained on the entire dataset) is trained for 150 epochs whereas the prompts in the APT method are trained for 80 epochs on their respective sources.  We emphasize that the paragon is never trained for fewer epochs than the APT method to remain a true ``upper bound".  For the paragon prompt tuning numbers we do not use structured attention.  Consistent with \cite{learnablememory} we use $5$ memory tokens at each layer for deep prompting.  For the prompt tuning of the APT method structured attention is applied during both train and inference time.  In all cases we optimize using the AdamW algorithm \cite{loshchilov2018decoupled} with the weight decay parameter set to $0.02$.  We use linear warmup cosine annealing with start learning rate $1\mathrm{e}{-5}$, minimum learning rate $1\mathrm{e}{-6}$, and one warmup epoch.  The base learning rates for finetuning, head-only finetuning, and prompt tuning are $1\mathrm{e}{-5}$, $5\mathrm{e}{-1}$, and $1\mathrm{e}{-1}$ respectively.  We did not do any hyperparameter sweep over learning rates.  The one exception is for the ``Bias+Head" column in Table~\ref{table:pt-vs-ft} we did a sweep over learning rates to arrive at the learning rate $5\mathrm{e}{-3}$.  However, we note that this column is merely for comparison and does not concern our specific method.  We use a batch size of $8$ and follow the convention presented in \cite{goyal2017accurate} of rescaling the learning rate by the effective batch size (batch size x devices x nodes) divided by $256$.
\par
We perform data augmentation following standard practice in training ViTs and include RandAugment \cite{NEURIPS2020_d85b63ef} with N=2 and M=10.  However we note that we did not use Mixup \cite{zhang2018mixup} which is known to be a reliable way of increasing performance.
\section{Dataset Details}
In Table~\ref{tab:suppl_datasets} we report detailed statistics for the datasets used as well as links to download the datasets.

\begin{table*}[!t]
\caption{\textbf{Dataset sample/class counts.} We list the number of training images, test images, and classes for each of the datasets.  We also provide a link to download the data.}
\resizebox{\textwidth}{!}{
\begin{tabular}{lcccl}
\toprule
     {\bf Dataset} & {\bf Training Images} & {\bf Testing Images} & {\bf \# Classes} & {\bf URL} \\
\hline
    MIT-67~\cite{mit67recognizing} & 5360 & 1340 & 67 & \footnotesize{\url{https://web.mit.edu/torralba/www/indoor.html}}\\
    CUB-200~\cite{WahCUB_200_2011} & 5994 & 5794 & 200 & \footnotesize{\url{https://www.vision.caltech.edu/datasets/cub_200_2011/}} \\
    Caltech-256 \cite{griffin_holub_perona_2022} & 15418 & 15189 & 257 & \footnotesize{\url{https://authors.library.caltech.edu/7694/}}\\
    Oxford Pets~\cite{parkhi12a} & 3680 & 3669 & 37 & \footnotesize{\url{https://www.robots.ox.ac.uk/~vgg/data/pets/}}\\
    FGVC-Aircrafts~\cite{maji13fine-grained} & 6667 & 3333 & 100 & \footnotesize{\url{https://www.robots.ox.ac.uk/~vgg/data/fgvc-aircraft/}}\\
    Oxford Flowers~\cite{Nilsback06} & 2040 & 6149 & 102 & \footnotesize{\url{https://www.robots.ox.ac.uk/~vgg/data/flowers/102/}} \\
    Stanford Cars~\cite{KrauseStarkDengFei-Fei_3DRR2013} & 8144 & 8041 & 196 & \footnotesize{\url{https://ai.stanford.edu/~jkrause/cars/car_dataset.html}}\\
    CIFAR-100 \cite{krizhevsky2009learning} & 50,000 & 10,000 & 100 & \footnotesize{\url{https://www.cs.toronto.edu/~kriz/cifar.html}} \\
    CORe50 \cite{pmlr-v78-lomonaco17a,Lomonaco2019FineGrainedCL} & 119,894 & 44,972 & 50 & \footnotesize{\url{https://vlomonaco.github.io/core50/}}\\ 
\bottomrule
\end{tabular}}
\label{tab:suppl_datasets}
\end{table*}

\section{Additional Ablations}
\begin{table*}[h]\small
\centering
\begin{tabular}{c|ccccccc} \toprule
    {Dataset} & {2 Shards} & {3 Shards} & {5 Shards} & {10 Shards} & {15 Shards} & {20 Shards} & {50 Shards} \\ \midrule
    MIT-67  & 3.1\% & 0.9\% & 0.6\% & 0.7\% & 0.6\% & 0.9\% & 0.5\%\\
    Cub-200 & 3.3\% & 1.1\% & 1.5\% & 0.8\% & 1.0\% & 0.8\% & 1.3\% \\
    Caltech-256 & 3.0\% & 1.4\% & 0.8\% & 0.2\% & 0.6\% & 0.4\% & 0.2\% \\
    Pets & 1.4\% & 0.2\% & 0.3\% & 0.0\% & -0.1\% & 0.2\% & 0.3\% \\ 
    Aircrafts & 6.2\% & 5.2\% & 5.0\% & 3.8\% & 3.2\% & 3.7\% & 3.0\% \\
    Flowers & 0.7\% & 4.2\% & 0.2\% & 0.6\% & 0.7\% & 1.3\% & 2.8\% \\
    Stanford Cars & 9.0\% & 7.6\% & 5.6\% & 5.5\% & 4.9\% & 5.1\% & 4.7\% \\ \midrule
     Average & 3.81\% & 2.94\% & 2.0\% & 1.66\% & 1.56\% & 1.77\% & 1.83\%
\end{tabular}
\caption{\textbf{Average vs. majority vote.} We report the accuracy of average ensembling minus the accuracy of majority vote.  We observe that average ensembling uniformly outperforms majority vote.}
\label{table:avg_vs_maj_vote}
\end{table*}
\paragraph{Average ensembling vs. majority vote.}
In our APT method we chose to aggregate the individual predictions of the prompts by average ensembling.  Another common ensembling method is to perform majority vote.  Consistent with \cite{machineunlearning} we find that average ensembling outperforms majority vote.  In Table~\ref{table:avg_vs_maj_vote} we report the performance gap of average ensembling over majority vote for the sharding experiment.  We see that excluding one exceptional case, average ensembling uniformly outperforms majority vote for all datasets and all numbers of shards.  The performance gain on average is in the range 1.5-3.8\%.

\paragraph{Pretraining.} To investigate how APT performs when the backbone transformer has a different pretraining, instead of using ImageNet21k we experiment with loading the VIT-B/16 from the visual encoder of the multimodal model ALBEF \cite{li2021align}.  In Table~\ref{table:shard_albef} we report the accuracies of APT applied to this checkpoint.  We see that the performance of APT for the ALBEF visual encoder decays more quickly as the number of shards increases relative to the ImageNet21k numbers reported in Table~\ref{table:shard_accs}.  For example for the visual encoder of ALBEF, for $10$ shards only the datasets MIT-67, Caltech-256, and Pets are within $5\%$ performance of the paragon, whereas by contrast for the ImageNet21k checkpoint all datasets except for Aircrafts and Stanford Cars are within $5\%$ performance of paragon even when the number of shards is twice as large, namely $20$.  Thus we conclude that the pretraining of the backbone transformer is highly pertinent for the performance of APT.  This is sensible as due to the structured attention the APT prompts do not modify the internal representations of the backbone, and thus are unable to provide compensation whenever the backbone representations are deficient.

\begin{table*}[h]\small
\centering
\begin{tabular}{c|c|ccccccc} \toprule
    {Dataset} & {No Sharding} & {2 Shards} & {3 Shards} & {5 Shards} & {10 Shards} & {15 Shards} & {20 Shards} & {50 Shards} \\ \midrule
    MIT-67 & 89.1\% & 87.5\% & 88.4\% & 88.9\% & 89.0\% & 88.4\% & 88.4\% & 87.3\% \\
    Cub-200 & 78.8\% & 71.6\% & 66.9\% & 57.1\% & 54.9\% & 48.5\% & 46.2\% & 39.6\% \\
    Caltech-256 & 91.4\% & 88.8\% & 89.2\% & 88.7\% & 88.9\% & 88.6\% & 87.8\% & 86.2\% \\
    Pets & 91.1\% & 89.9\% & 88.2\% & 87.0\% & 86.3\% & 83.1\% & 81.0\% & 66.1\% \\ 
    Aircrafts & 72.6\% & 60.5\% & 54.4\% & 51.1\% & 40.2\% & 38.7\% & 35.9\% & 30.7\% \\
    Flowers & 93.4\% & 85.1\% & 83.1\% & 81.7\% & 80.7\% & 78.1\% & 76.2\% & 67.8\%  \\
    Stanford Cars & 83.3\% & 78.2\% & 76.5\% & 70.7\% & 63.7\% & 59.5\% & 55.9\% & 43.6\%  \\ \midrule
     Average & 85.67\% & 80.23\% & 78.1\% & 75.03\% & 71.96\% & 69.27\% & 67.34\% & 60.19\%
\end{tabular}
\caption{\textbf{Sharding from ALBEF pretraining.} We report the accuracies for the sharding experiment using the ALBEF checkpoint.}
\label{table:shard_albef}
\end{table*}

\paragraph{Finetuning.} While inference and storage for the APT method is less costly than ensembling finetuned models, it is worthwhile to ask how the two compare in terms of classification accuracy.  In Table~\ref{table:shard_ft} we report the accuracies for the sharding experiment using finetuning instead of APT.  Specifically we finetune separate models on each shard which are then ensembled at inference time.  By comparing the results in Table~\ref{table:shard_accs} to the results in Table~\ref{table:shard_ft}, we see that APT uniformly outperforms finetuning in terms of classification accuracy, and the gap becomes most pronounced as the number of shards increases.  Specifically, for 20 and 50 shards APT has average accuracy of 77.3\% and 73.9\% respectively compared to 41.5\% and 25.6\% for finetuning.  We believe this is due to finetuning being more susceptible to overfitting when there are fewer data in contrast to APT which uses a fixed backbone and thus has a stronger inductive bias.

\begin{table*}\small
\centering
\begin{tabular}{c|c|ccccccc} \toprule
    {Dataset} & {No Sharding} & {2 Shards} & {3 Shards} & {5 Shards} & {10 Shards} & {15 Shards} & {20 Shards} & {50 Shards}  \\ \midrule
    MIT-67 & 87.1\% & 86.1\% & 83.8\% & 81.9\% & 74.4\% & 69.9\% & 68.8\% & 44.9\% \\
    Cub-200 & 88.4\% & 81.8\% & 76.4\% & 70.9\% & 54.4\% & 42.5\% & 32.5\% & 5.9\% \\
    Caltech-256 & 93.5\% & 90.3\% & 87.8\% & 85.8\% & 81.0\% & 78.2\% & 74.1\% & 52.0\% \\
    Pets & 94.5\% & 93.6\% & 92.6\% & 91.2\% & 89.7\% & 84.2\% & 81.6\% & 55.8\% \\
    Aircrafts & 75.6\% & 51.1\% & 44.5\% & 36.2\% & 24.1\% & 23.0\% & 19.2\% & 12.3\% \\
    Flowers & 97.4\% & 75.3\% & 56.1\% & 39.8\% & 15.6\% & 11.1\% & 2.2\% & 2.0\% \\
    Stanford Cars & 84.3\% & 53.3\% & 39.4\% & 28.2\% & 19.2\% & 16.2\% & 11.8\% & 6.4\% \\ \midrule
     Average & 88.69\% & 75.93\% & 68.66\% & 62.00\% & 51.20\% & 46.44\% & 41.46\% & 25.61\% \\
\end{tabular}
\caption{\textbf{Sharding using finetuning.} We report the accuracy for the sharding experiment when using finetuning.}
\label{table:shard_ft}
\end{table*}

\end{document}